%% file: main.tex
\documentclass[10pt,twocolumn,letterpaper]{article}

\usepackage{cvpr}

\usepackage{times}
\usepackage{epsfig}
\usepackage{graphicx}
\usepackage{amsmath}
\usepackage{amssymb}

\begin{document}

\title{Adaptive Deep Iris Feature Extractor at Arbitrary Resolutions}

\author{Yuho Shoji, Yuka Ogino, Takahiro Toizumi, Atsushi Ito\\
NEC Corporation\\
{\tt\small yuho-shoji@nec.com, yogino@nec.com, t-toizumi\_ct@nec.com, ito-atsushi@nec.com}
}

\maketitle

\begin{abstract}
This paper proposes a deep feature extractor for iris recognition at arbitrary resolutions. Resolution degradation reduces the recognition performance of deep learning models trained by high-resolution images. Using various-resolution images for training can improve the model's robustness while sacrificing recognition performance for high-resolution images. To achieve higher recognition performance at various resolutions, we propose a method of resolution-adaptive feature extraction with automatically switching networks. Our framework includes resolution expert modules specialized for different resolution degradations, including down-sampling and out-of-focus blurring. The framework automatically switches them depending on the degradation condition of an input image. Lower-resolution experts are trained by knowledge-distillation from the high-resolution expert in such a manner that both experts can extract common identity features. We applied our framework to three conventional neural network models. The experimental results show that our method enhances the recognition performance at low- resolution in the conventional methods and also maintains their performance at high-resolution.

\end{abstract}

\section{Introduction}
\label{sec:intro}
Iris recognition is one of the most reliable and discriminative methods for biometric identification. It is used in various fields, including forensic science, border control, and biometric payment systems. Iris recognition at high accuracy requires high-resolution (HR) iris images with fine textures. Here, resolution refers to the level of detail in the image \cite{ISO2023}. The International Organization for Standardization (ISO) recommends capturing an iris image at a rate of ten or more pixels per mm under in-focus conditions \cite {ISO2021}. However, high-resolution capturing causes optical difficulties: shallow depth-of-field and narrow field-of-view, both of which constrain the capturing environment. Due to these constraints, subjects need to get close to the camera to obtain an HR image of the iris, which is inconvenient for users.

\begin{figure}[t]
  \begin{center}
   \includegraphics[width=0.95\linewidth]{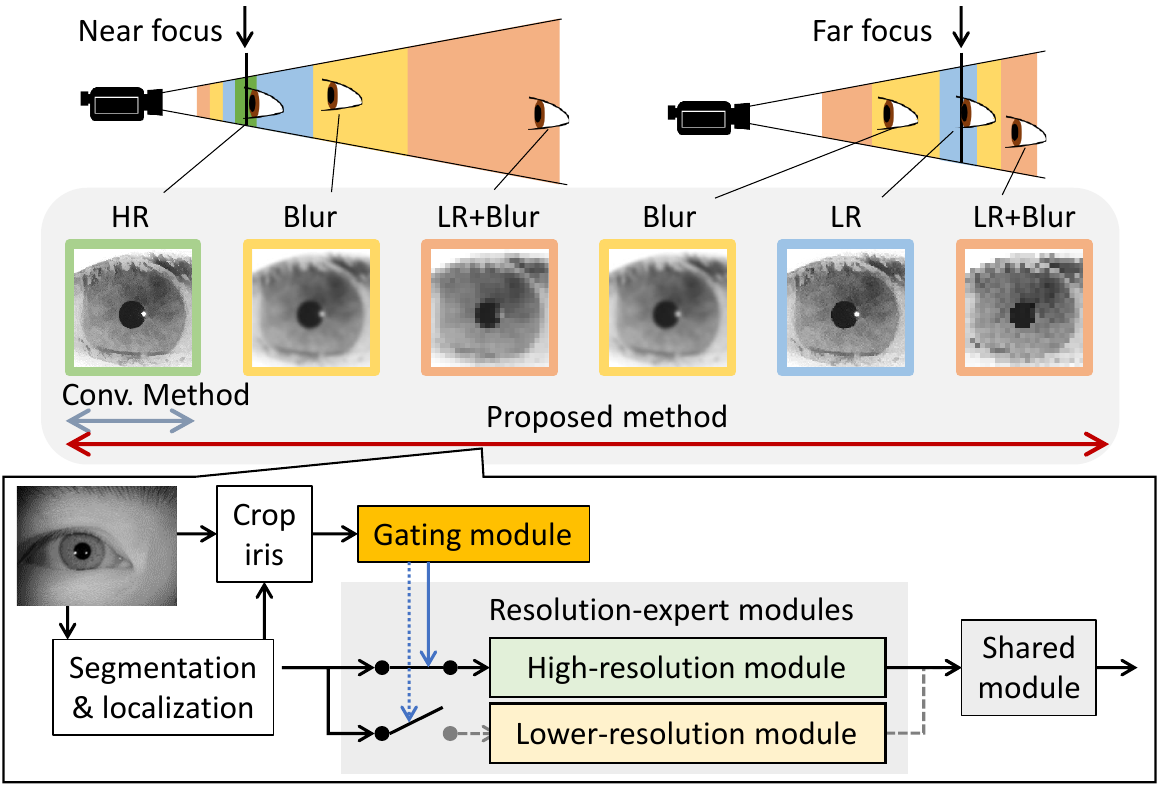}
  \end{center}
  \caption{Proposed iris recognition method using automatically switching networks for iris images with arbitrary resolutions.}
  \label{fig:snap-shot}
 \end{figure}

To develop a user-friendly iris recognition system, we need to achieve iris recognition in less constrained capturing environments. In such environments, the resolution of an iris image can be degraded under various conditions depending on the target eye positioning, as shown in the upper side of Figure \ref{fig:snap-shot}. This resolution-degradation is caused by both optical down-sampling and out-of-focus blurring. These two factors occur simultaneously, and separating them on the basis of resolution degradation is difficult. Thus, the user-friendly system requires an iris recognition engine that is robust to arbitrary resolution degradation.

We focus on a deep learning-based method because of its extensibility and potential for improving degradation robustness. Several deep learning-based iris recognition methods have been explored to extract robust features to image quality degradation. \cite{boutros2022low, chen2020tcenter, kawakami2022simple, wang2022d, yang2021dualsanet}. (Note that non-deep-learning methods are out of the scope of this paper because their encoders are generated in a hand-crafted or unsupervised manner.) Some conventional methods have introduced metric learning to deal with unknown classes for recognition. These methods train models using HR iris images. They show high recognition performance at HR while their performance drops at low-resolution (LR). Conventional studies have proposed image-restoration-based methods and data-augmentation-based methods to enhance the recognition performance at LR. The image-restoration-based methods use deblurring \cite{jamaludin2021deblurring,lee2021enhanced} or super-resolution \cite{IrisDNet,wang2022d} as a pre-processing step. While these methods reconstruct HR images from LR ones, they necessitate additional networks and impose an extra computational cost. Data-augmentation-based methods train models with down-sampled LR images \cite{boutros2022low}. This method improves the recognition performance at LR without additional computation. However, using lower-resolution images for model training degrades the recognition performance at HR due to the limit of capacity for generalization. Thus, a method is required to solve both the computational cost and generalization.

We propose a resolution-adaptive iris feature extractor that meets these requirements. Our method uses an automatically switching network scheme \cite{Cai_2021_WACV, eigen2013learning, mullapudi2018hydranets, wang2020deep}. The method includes three kinds of modules: a gating module, resolution-expert modules (REMs), and a shared module, as shown in the lower side of Figure \ref{fig:snap-shot}. The gating module automatically switches REMs based on the resolution of an input image. The REMs have three resolution expert networks. Each expert is trained to specialize in different resolution degradation. The difficulty of applying this scheme to an iris recognition task is that different experts must extract common identity features from different input images. To overcome this difficulty, our method trains lower-resolution experts using a manner of knowledge distillation from the HR expert. Furthermore, the method newly introduces the shared module on the output side to enable more similar features to be extracted from each expert. We applied our framework to three conventional deep learning-based methods, and the experimental results show that our method achieves better recognition performance than the baseline methods under various degradations. We also show that the additional runtime in applying our method is sufficiently lower than that of baseline feature extraction methods and image-restoration-based methods.

 \section{Related Work}
To achieve iris recognition in less-constrained capturing environment, conventional studies have proposed hardware-based methods and software-based methods. Some hardware-based methods use imaging sensor that can adjust its focus \cite{IOTM2006, zhang2020all}. While they can capture in-focus iris images, they require more complex and expensive camera systems. Software-based methods, on the other hand, are not constrained by such hardware limitations, enabling simple and cost-effective system development. It also offers greater flexibility in terms of camera system selections and capture environments compared to hardware replacements. We focus on a software solution in this paper for these reasons. 
 
Iris recognition is based on the following processing: segmentation, localization, iris image normalization, feature extraction and matching. Among these processes, image quality degradation directly affects the performance of the feature extraction processes. The segmentation and localization methods extract global edge features of an iris or pupil boundary. These features are maintained even in small iris so that these methods can correctly detect iris regions from LR images \cite{arsalan2018irisdensenet, Toizumi_2023_WACV, zhao2021detection}. On the other hand, in the feature extraction process, both local and global iris textures are extracted. LR iris images lose detail in iris texture, making it difficult to extract the same feature from an LR iris image as from the HR one. Some recent works focused on resolution-robust feature extraction \cite{boutros2022low, IrisDNet,Ribeiro2019b}.

Conventional iris feature extraction methods encode normalized iris images to feature templates using hand-crafted methods \cite{czajka2019domain, daugman2009iris, masek2003recognition, miyazawa2008effective}. They compare templates and use spatial iris features for recognition. Recent studies have introduced deep learning to iris recognition because they have a potential for improving recognition performance and robustness to low-quality images \cite{alinia2022boosting,chen2020tcenter,jalilian2022offangle,kawakami2022simple,wei2022towards, yang2021dualsanet,zhao2017towards}. Some methods train deep learning models in metric-learning scheme. The models can extract similarity-comparable features from an input image and deal with unknown classes. Some studies have utilized data augmentation methods for training the models to extract degradation-robust features \cite{boutros2022low, kawakami2022simple}. As another approach for robust feature extraction, uncertainty embedding (UE) has been proposed to estimate uncertainty and identity features as independent components\cite{wei2022towards}. Simple and accurate convolutional neural network  (SACNN) calculates the occlusion-aware weighted matching score and enhances occlusion-robustness \cite{kawakami2022simple}. This paper focuses on metric learning methods because they can discriminate unknown classes and have the potential of higher robustness and scalability.

  \begin{figure*}[t]
  \begin{center}
   \includegraphics[width=0.9\linewidth]{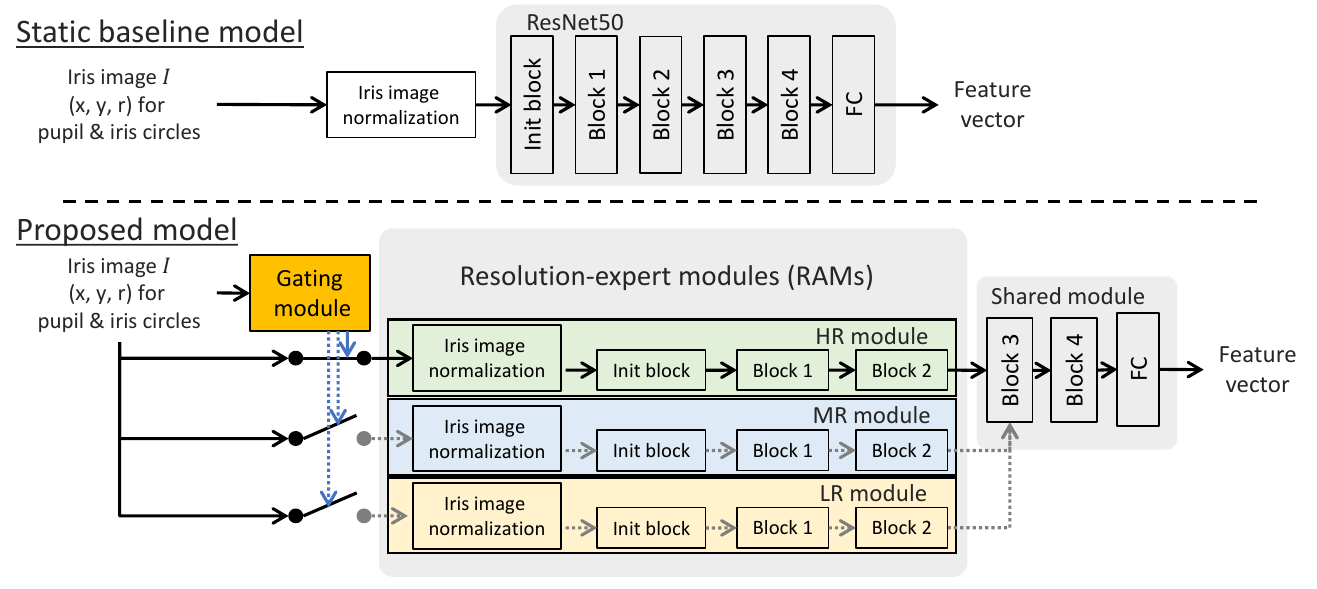}
  \end{center}
  \caption{Architecture difference between conventional baseline and proposed models. The proposed feature extractor has resolution-expert modules (REMs) on the input side and shared module on the output side. Each expert in REMs is trained for different ranges of resolution. Our gating module selects a module in REMs for feature extraction from a resolution condition of an input image.}
  \label{fig:architecture}
 \end{figure*}

For improving the robustness to degraded images, loss function and knowledge distillation have been proposed for training metric learning models. Some methods use L2-normalized feature embedding to exclude degradation features from identity features \cite{Deng_2019_arcface,Meng_2021_CVPR,ranjan2017l2,wang2018cosface}. Ring loss improves robustness by constraining the L2-norm of the features \cite{zheng2018ring}. Knowledge distillation or knowledge transfer \cite{hinton2015distilling,li2017learning} train a feature extractor for degraded images using features extracted from high-quality images \cite{boutros2022low,ge2018low,miyamoto2021joint}. A model trained by distillation can enhance recognition performance for LR, while maintaining a computational cost. However, since there is a trade-off between robustness to resolution-degradation and recognition performance for HR, these training methods reduce the performance for higher resolution images. 

Super-resolution (SR) methods train an encoder-decoder network to extract high-quality images from degraded images \cite{Chen2020IASR,IrisDNet,Ribeiro2019b,Ribeiro2017,wang2022d}. IASR and IrisDNet train an SR network with a feature extractor to keep discriminative features \cite{Chen2020IASR,IrisDNet}. Since they fix the parameters of the feature extractor, performance for high-quality images is maintained. However, the SR model imposes a high computational cost due to its additional network. Also, conventional iris image SR methods are designed to upscale with a single and fixed magnification rate. We need a method that can accurately recognize iris images with arbitrary resolution while maintaining a lower computational cost. 

Dynamic neural networks are one candidate for achieving higher recognition performance and lower computational cost \cite{han2022dynamic}. Methods using such networks dynamically estimate the network structure, including network modules, shortcut layers, depths, input sizes, and weight parameters from input images. Among the various types, we focus on mixtures of experts (MoE) \cite{Ebrahimpour2011, eigen2013learning,ma2018modeling}, which have multiple network branches as experts in parallel. Some MoEs methods introduce a hard gating module that selects an execution branch among multiple experts from an input image for improving computational efficiency \cite{Cai_2021_WACV,mullapudi2018hydranets,wang2020deep}. We utilize this automatically switching network with hard gating to enhance the robustness to resolution degradation while achieving lower computational cost. Our model has three resolution expert branches, and a gating module selects them based on the resolution condition of the input image. In general, different modules extract different features, while each expert must extract common identity features for iris recognition. To realize this, we train lower-resolution experts in a distillation manner and design network architecture with a shared module on the output side. To the best of our knowledge, this is the first work to introduce MoEs with hard gating to deep-learning-based iris feature extraction.
 
 \section{Proposed Method}
Our proposed framework utilizes an automatically switching network to achieve higher recognition performance for various resolutions. The proposed framework has three modules: a gating module, resolution-expert modules (REMs) and a shared module, as shown in Figure \ref{fig:architecture}. REMs include three modules, and each module is specialized for different ranges of resolutions. The gating module first selects a module in REMs based on the resolution of an input image. Then, the selected module in REMs and the shared module extract a feature vector from an iris image. Our distillation-based training scheme enables common identity feature extraction using each expert module. The proposed framework requires a lower additional computational cost. Our method based on ResNet50 takes 73.0 ms, which is only 5.1 ms slower than the baseline model for a single-image feature extraction on a CPU. Detailed runtime comparison is provided in the the supplementary material.

\begin{figure}[t]
 \centering
 \includegraphics[width=0.99\linewidth]{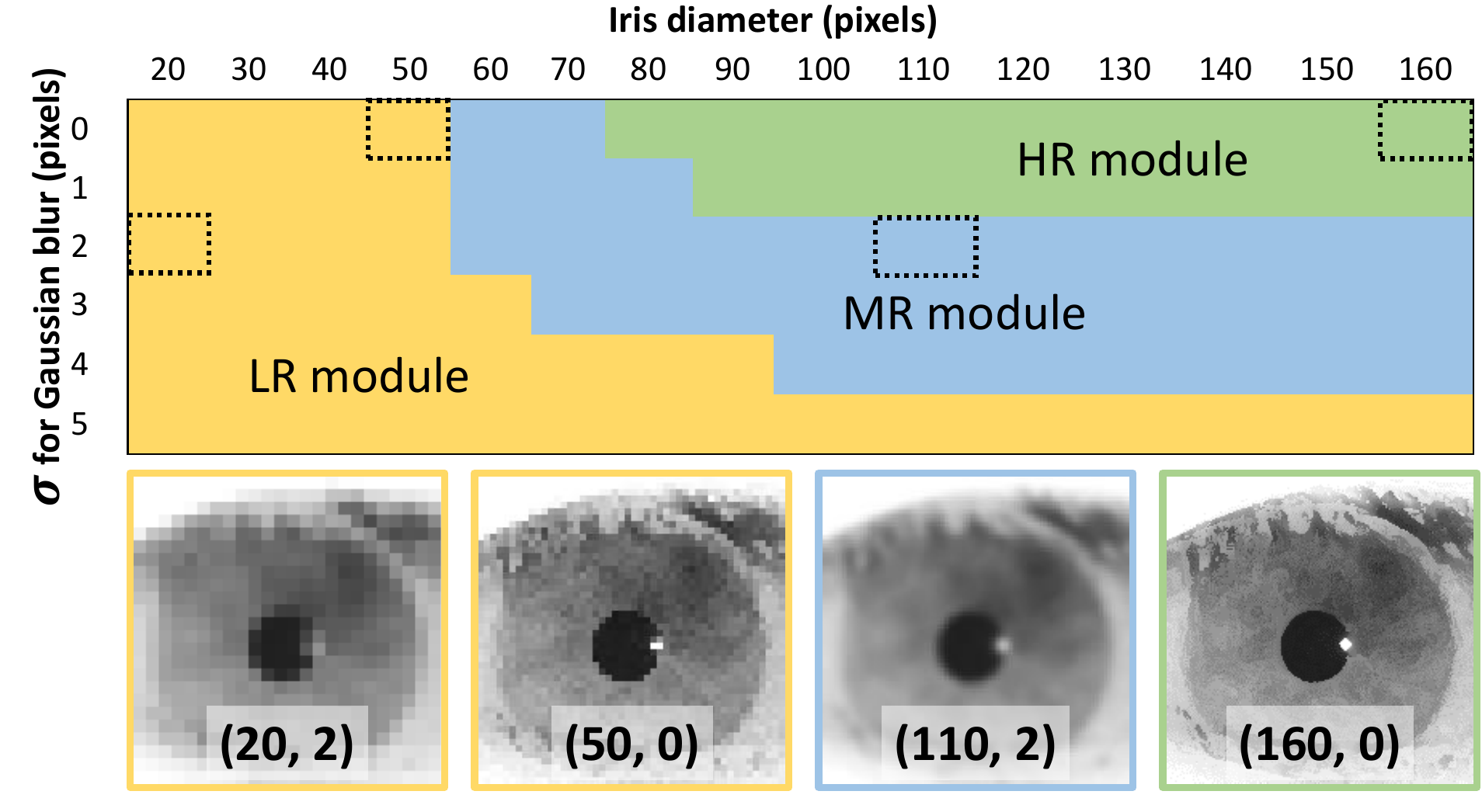}
 \caption{Best-performing module labels for each degradation condition (upper) and examples of degraded iris images (lower). }
 \label{fig:gate-label}
\end{figure}

\subsection{Architecture of modules}
\label{subseq:architecture}

\begin{figure}[t]
 \begin{center}
  \includegraphics[width=0.96\linewidth]{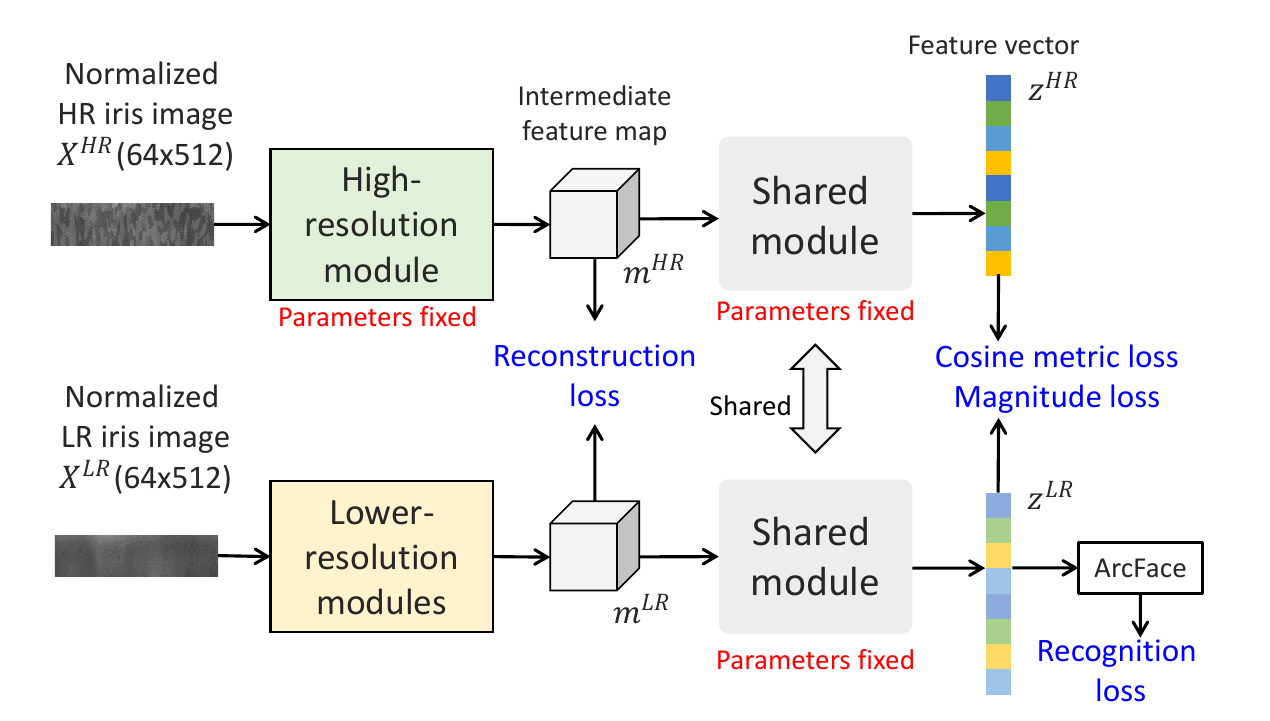}
 \end{center}
 \caption{Training scheme for proposed lower-resolution modules. First, HR module and shared modules are trained for iris recognition, Afterward, lower-resolution module is trained with four losses while parameters of HR and shared modules are fixed.}
 \label{fig:training}
\end{figure}

We can use conventional deep-learning models as a backbone for feature extraction. We choose three conventional models: ResNet50, SE-ResNet18 for simple and accurate CNN (SACNN) \cite{kawakami2022simple} and Max-out4 for uncertainty-guided curriculum learning (UE+UGCL) \cite{wei2022towards}. Among these, ResNet50 has been used in several recent iris recognition and low-resolution iris recognition works \cite{Ahmad2019BTAS,boutros2022low,Boyd2019BTAS}. Here, we explain the architecture of our method based on ResNet50. As shown in the upper side of Figure \ref{fig:architecture}, ResNet50 has an init block, four intermediate blocks and a fully-connected layer. Our feature extractor and baseline static model inputs iris images normalized using a rubber sheet model \cite{Daugman2007}. The size of the normalized iris image was $64\times 512$, the same as in the conventional method \cite{Othman2016OSIRIS}. The dimension of a feature vector is set to 256 according to \cite{boutros2022low}. 

The bottom of Figure \ref{fig:architecture} shows the detailed architecture of the proposed models. The proposed method splits the feature extractor into REMs on the input side and the shared-module on the output side. The REMs have three expert modules: high-resolution (HR) module, middle-resolution (MR) module and low-resolution (LR) module. Each module is trained to be specialized for different levels of degradation. Since the MR-module and LR-module extract lower-resolution images than the HR module, we call them lower-resolution modules in the following sections. The shared module on the output side guides extracting comparable features using different REMs. We empirically split REMs and the shared module at the point between blocks 2 and 3 to enhance robustness. The split position is evaluated in an ablation study in the experiments section.

The gating module extracts a module label of REMs from the iris image degraded both down-sampling and blurring. As a down-sampling condition, we can utilize the pixels of the iris diameter extracted through iris localization process. Then, estimating the blur level enables the selection of a module in REMs conditioned on two degradation factors. However, robustly estimating the blur level is a difficult task under various down-sampling conditions. Then, our gating module estimates a module label in REMs directly from an input image in a way of the classification. Three-class classification is a simpler task than estimating the degradation level, allowing the gating module to work with a lightweight network. The architecture of the gating module is constructed based on the shallow parts of ResNet50. The details of the architecture are provided in the supplementary material. Its input is a cropped square iris image resized to 192 pixels with an iris diameter of 160 pixels. The output is the expert label (HR, MR, and LR) indicating which expert model is best suited for the input. 

\subsection{Training method}
\label{subseq:loss}

Figure \ref{fig:training} shows the proposed feature extraction training scheme. First, the HR and shared modules are trained with HR images for iris recognition. The trained network parameters are then fixed, and the MR and LR  modules are independently trained with lower-resolution images using a distillation manner. The degradation levels of the training images depend on each lower-resolution module: the lower the resolution of a module, the greater the degradation.

When a lower-resolution module is trained, a clean image and a degraded image are fed into the HR module and lower-resolution module, respectively. Each lower-resolution module is trained so that the features from the LR image are close to the features from the HR image in both the intermediate feature maps and output feature vectors. This training scheme utilizes knowledge in the feature space of a model trained by clean HR images and trains a lower-resolution module to extract robust, discriminative and comparable features.

The lower-resolution modules are optimized using four loss functions: classification loss, reconstruction loss, cosine metric loss and magnitude loss. The latter three loss functions are inspired by super-resolution methods for face recognition \cite{Chen2020IASR}, which simultaneously super-resolves in image space and feature vector space. As a classification loss, we use a large margin cosine loss (ArcFace) $L_{arc}$ to enhance the discrimination performance \cite{Deng_2019_arcface}. ArcFace has trainable parameters that are representative of each class in the feature vector space. These parameters are optimized when training the shared module, and they are fixed when training the lower-resolution modules.
Reconstruction loss $L_r$ is used for stable training and boosting recognition performance. It is defined as the L2 distance between the intermediate feature maps for HR $m^{HR}$ and for lower-resolution $m^{LR}$ extracted by a lower-resolution module. 
\begin{equation}
	L_r = \frac{1}{n}\sum_{i=1}^{n}\left|| m^{LR}_i-m^{HR}_i|\right|^2_2,
	\label{eq:loss_r}
\end{equation}
where $||\cdot||_2$, $n$ and $i$ denote L2 norm, batch size and batch index, respectively.

In addition to the reconstruction loss, we use a cosine metric loss $L_{cos}$ and magnitude loss $L_{mag}$ \cite{Chen2020IASR,IrisDNet}. These losses make lower-resolution feature vectors $z^{LR}$ close to the HR feature vector $z^{HR}$. They contribute to the recognition performance and robustness. The cosine metric loss $L_{cos}$ is defined as the cosine similarity between $z^{LR}$ and $z^{HR}$, and it is described as:
\begin{equation}
	L_{cos} = \frac{1}{n}\sum_{i=1}^{n}\left(1-\frac{\left(z_i^{LR}\right)^T z_i^{HR}}{||z_i^{LR}||_2||z_i^{HR}||_2} \right).
	\label{eq:loss_cd}
\end{equation}

The magnitude loss $L_{mag}$ is defined as the L1-distance between the L2 norm of $z^{HR}$ and $z^{LR}$,
\begin{equation}
	L_{mag} = \frac{1}{n}\sum_{i=1}^{n}\left|\left| ||z_i^{LR}||_2-||z_i^{HR}||_2 \right|\right|_1.
\end{equation}

The total loss for training the lower-resolution modules is defined as the weighted sum of the four losses:
\begin{equation}
	\label{eq:loss_total}
	L_{total} =L_{arc} + \alpha_{r}L_r + \alpha_{cos}L_{cos} + \alpha_{mag}L_{mag},
\end{equation}
where, $\alpha_r$, $\alpha_{cos}$ and $\alpha_{mag}$ are weights for each loss. In this experiment, we set $\alpha_{r}$, $\alpha_{cos}$ and $\alpha_{mag}$ to 0.1, 1, 0.0001 and 1, respectively, referenced to the facial image SR \cite{Chen2020IASR}.

We trained the gating module to estimate the best expert label from an input image after training each expert module and the shared module.  Firstly, we investigated the experts with the best EER for each degradation condition. Figure \ref{fig:gate-label} shows the best expert label for each degradation condition. Iris images for training were randomly degraded and and gating module was trained to classify the best expert label from the image using cross-entropy loss. 

As summarizing our network design and training scheme, we can transform a static baseline model into our method by adding lower-resolution modules and the gating module. Because the architecture and weights of parameters of a baseline model are maintained, we can easily introduce our dynamic scheme to conventional deep-learning models \cite{kawakami2022simple,wei2022towards}. Also, our method easily enhances the robustness of target degradation ranges by adding a lower-resolution module trained by the target ranges.

\section{Experiments}
We evaluated our method through five experiments: two ablation studies and three comparisons. First, we investigated the split point of a network for boosting the robustness for LR images. Next, the effectiveness of the gating module was evaluated using a dataset containing blurred images. Then, we compared the recognition performance of the proposed method with conventional methods for each down-sampling condition and blurring condition, respectively. Then, we evaluated robustness to various and unknown degradation conditions using iris images degraded due to both down-sampling and blurring. We used ResNet50 as a baseline model for these four experiments. Finally, we apply our method to two other baseline models and compare our method and conventional methods.

\subsection{Datasets}
We used four public datasets: CASIA-Iris-Thousand, CASIA-Iris-Lamp, CASIA-Iris-Distance and MMU2 \cite{mmudataset,casia}. CASIA-Iris-Thousand has 20,000 images of 1,000 subjects. Images of the first 800 subjects were used for training, and images of the remaining 200 subjects were used for testing. CASIA-Iris-Lamp has 15,512 images of 394 subjects. CASIA-Iris-Distance has 5,134 images of 142 subjects. MMU2 has 995 images of 100 subjects. The CASIA-Iris-Lamp, CASIA-Iris-Distance and MMU2 datasets were used for testing in a generalization evaluation. We localized the center coordinates and radii of the pupil and iris circles by using iris localization networks \cite{Toizumi_2023_WACV} for all four datasets. The square region around the iris was cropped and resized to $192 \times 192$ image sizes with an iris diameter of 160 pixels. For all resize processing, we used area interpolation to emulate the optical sampling on image sensors. These $192 \times 192$ images were treated as clean HR images.

\subsection{Training condition} 
We first trained the HR and shared modules with HR iris images and then trained MR and LR modules independently. We trained the HR and shared modules using ArcFace loss \cite{Deng_2019_arcface}. The value of the scale parameter was set to 30 for stable training because the default value of 64 \cite{Deng_2019_arcface} caused unstable convergence and gradient vanishing. The initial value of the margin was set to 0, and after 2k iteration, we set the margin to 0.45, which is a value proposed in \cite{boutros2022low}. This margin-scheduling method allows stable convergence of loss function. The optimization method was stochastic gradient descent (SGD) with an initial learning rate of 0.1, a momentum value of 0.9, and a weight decay of 0.0005. These parameters were used to train conventional metric learning models \cite{boutros2022low,Deng_2019_arcface,wei2022towards}. The batch size was set to 64. The total number of iterations was 30k, and the learning rate was multiplied by 0.1 for both 15k and 27k iterations. After training the HR and shared modules, we fixed their parameters and trained the MR and LR modules using the total loss (Equation \ref{eq:loss_total}). The initial learning rate of SGD was 0.01, and the weight decay was 0.0005. The batch size was set to 64. The number of iterations was set to 30k, and the learning rate was multiplied by 0.1 at 21k iterations.

As data augmentation, We applied random-horizontal shift and brightness-contrast to all training images for feature extraction models. They were used for training robust CNN-based iris recognition models \cite{kawakami2022simple,Xu2017}. The range of random horizontal shift was 10 pixels for both the left and right sides. The ranges for both brightness and contrast variation were $\pm$ 50\%. To train each REMs module, we also applied Gaussian blur (GB) and iris-diameter-based down-sampling. Different image degradation ranges were applied to train each resolution-expert module, and degradation levels were sampled uniformly with the ranges assigned for the training. The ranges of the standard deviation ($\sigma$) for the Gaussian blur were (0.0, 1.0), (1.0, 3.0) and (3.0, 5.0) for the HR, MR and LR modules, respectively. The ranges of the down-sampling iris diameter were (120, 160), (60, 120) and (20, 60) for the HR, MR and LR modules, respectively. We applied these augmentations in the order of brightness-contrast, random-horizontal shift, Gaussian-blurring and down-sampling. Each data-augmentation process was applied with a probability of 0.5.

We trained the gating module to estimate the best expert label from an input image by using cross-entropy loss. Training iris images are randomly degraded by both down-sampling and Gaussian-blurring, and the expert label for each image is decided based on the results of Figure \ref{fig:gate-label}. We used SGD as an optimizer with the learning rate 0.001. The batch size was 128, and the number of iterations was 40k.

\subsection{Comparison methods}
We compared the proposed method with baseline models trained in four ways and three image-restoration-based methods. Here, we describe each model where ResNet50 is used as a baseline model. We trained ResNet50 using four training methods: ArcFace \cite{Deng_2019_arcface} with only HR images (Res50-HR), ArcFace \cite{Deng_2019_arcface} with LR images (Res50-LR), Ring loss \cite{zheng2018ring} with LR images (Res50-Ring loss) and knowledge transfer (KT) \cite{boutros2022low} with LR images (Res50-KT). Ring loss constrains feature norms to learn resolution-robust features. KT method transfers knowledge of feature representations from the Res50-HR to the model trained with LR iris images. KT is a state-of-the-art training method for LR iris recognition. For training of the Res50-LR, Res50-Ring loss, Res50-KT, we applied both down-sampling and Gaussian-blurring as data augmentation for fair comparison. The range of the standard deviation ($\sigma$) for GB was set to (0, 5), and the range of the iris diameter was set to (20, 160). The optimization settings of baseline models were the same as those for the HR and shared modules.

As image-restoration-based methods, we compared IrisDNet \cite{IrisDNet}, D-ESRGAN \cite{wang2022d} and DeblurGAN \cite{kupyn2018deblurgan, lee2021enhanced}. D-ESRGAN integrate CNN and vision transformer \cite{dosovitskiy2020image} for iris image SR to restore local and global feature of iris images. Because IrisDNet and D-ESRGAN upscale iris images with a fixed magnification rate, we evaluate their performance for iris images with only their target iris diameters. DeblurGAN restores clean images from blurred images using U-Net architecture. The training settings of these methods are described in the supplemental material. We used Res50-HR as a baseline feature extractor. 

\subsection{Metrics}
We used equal error rate (EER) to evaluate the recognition performance. EER is an error rate where the false rejection and false acceptance rates are equal. It is used in previous LR studies \cite{boutros2022low,IrisDNet,Ribeiro2019b}. For more detailed comparison, we used d-prime and detection error trade-off (DET) curve and provide the results in the supplementary material. We calculated the cosine similarity between features from HR images and those from LR images and used it as a matching score.

\begin{figure}[t]
 \centering
  \includegraphics[width=0.99\linewidth]{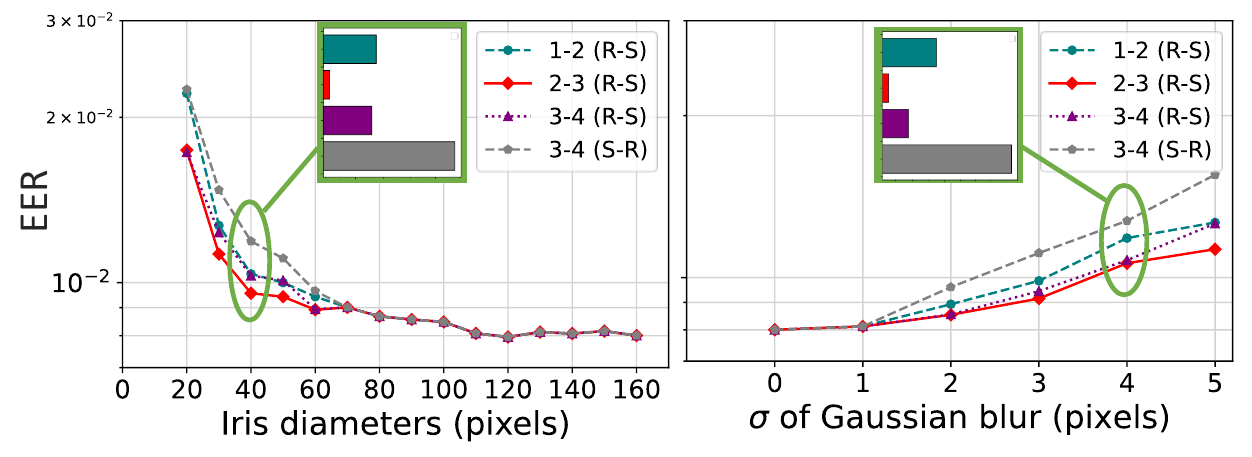}
 \caption{Results (EER) of ablation study using CASIA-Iris-Thousand for split points between REMs and shared module. Bold and underlined mean the best and second, respectively.}
 \label{fig:ablation}
\end{figure}

\subsection{Ablation study}

\begin{table}[t]
  \caption{Results (EER) of ablation study for gating module using CASIA-Iris-Thousand.}
  \small
   \label{table:exp_gate}
   \begin{center}
    \begin{tabular}{lccc}\hline
     &\multicolumn{3}{c}{Blur degradation ranges} \\
     Method               &    (0, 1) &   (0, 3)  &    (0, 5)   \\ 
     \hline 
     HR-only             &    0.0343 &    0.0400 &    0.0555 \\
     Random gating        &    0.0215 &    0.0236 &    0.0336 \\
     Iris diameter gating &    0.0096 &    0.0100 &    0.0210 \\
     Gating module (Proposed)  &   \bf{0.0092}&   \bf{0.0091}&   \bf{0.0100}\\ 
     \hline
  \end{tabular}
   \end{center}
  \end{table}

  \begin{figure*}[t]
    \begin{center}
     \includegraphics[width=0.99\linewidth]{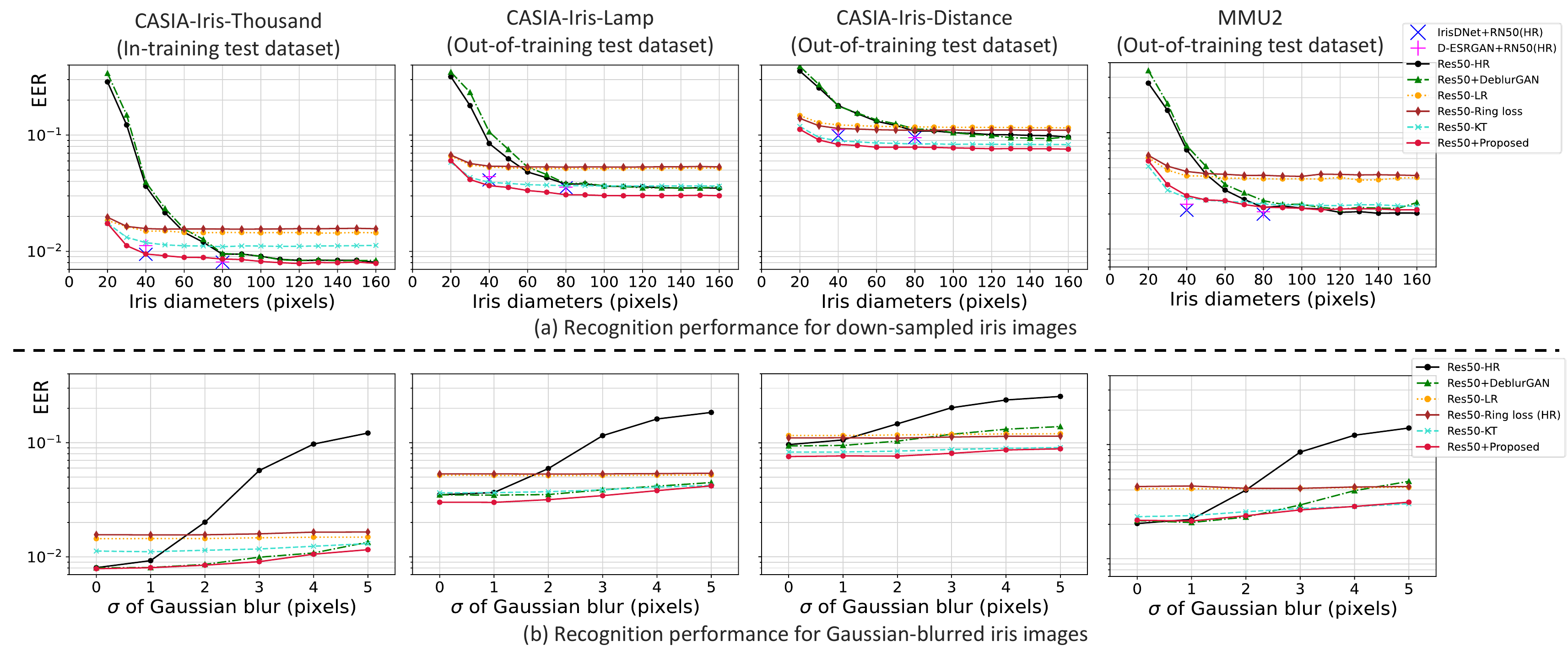}
    \end{center}
    \caption{Recognition performance comparison for down-sampled images (a) and Gaussian-blurred images (b). Upper side shows EER for down-sampled images without Gaussian blur. Lower side shows EER for Gaussian-blurred images with iris diameter of 160 pixels. Our proposed model and conventional models are trained using only CASIA-Iris-Thousand dataset.}
    \label{fig:compare}
   \end{figure*}
   
The performance of our framework largely depends on the split point between the REMs and shared module. To identify the optimal split point, we compared the recognition performance between different split points using the CASIA-Iris-Thousand dataset. Block 1-2 had a split point between block 1 and 2. In addition to the split point, we evaluated the order of the REMs and shared module in order to compare the performance with resolution adaptation in the shallow feature space with that in the deep feature space. R-S models had REMs on the input side and shared module on the output side. S-R models had shared module on the input side and then, network is split to REMs on the output side. We compared the performances of five architectures: block 1-2 (R-S), block 2-3 (R-S), block 3-4 (R-S) and block 3-4 (S-R). We used the EER for different down-sampling and blur conditions as the metric. For clean gallery images, we used the HR module in REMs and the shared module as the feature extractor. The probe images were degraded by down-sampling or blurring, and the EER was obtained for each module in REMs. The best EER for each expert was selected as the performance for that setting. Note that the gating module was not used for this evaluation.

  Figure \ref{fig:ablation} shows the results of the ablation study of the split point between the REMs and shared module. Block 1-2 (R-S) showed a better EER than block 3-4 (S-R) even though both have one intermediate block in REMs. This demonstrates that earlier adaptation is better than later for input images degraded by down-sampling or blurring. A comparison of the the split point of the REMs and the shared module showed that block 2-3 (R-S) had the best performance. Deeper REMs can expand the capacity of resolution adaptation. However, block 2-3 (R-S) had better performance than block 3-4 (R-S). This result indicates that the shallow shared module makes it difficult to extract comparable features using different resolution adaptive modules.

  In addition to the split point, we also evaluated the performance of the gating module using the CASIA-Iris-Thousand dataset. We compared the gating policy of our method with four other policies: HR-only, random gating, iris diameter gating and label gating. HR-only uses the HR-module regardless of the input image. Random gating randomly selects a module from the three modules. Iris diameter gating selects a module on the basis of the number of pixels in the iris diameter without assuming blur. Our proposed gating module estimates one module from resolution condition of an input image. We compared these policies using three blurring-level datasets to validate that the gating module can estimate which module is appropriate for blurred images. We generated degraded test datasets by applying down-sampling and Gaussian blurring operations, and the total number of images was 20,000. For three datasets, the range of the pixels of an iris diameter was set to (20, 160). The standard deviation ($\sigma$) of GB was randomly sampled from three ranges: (0, 1), (0, 3) and (0, 5).

Table \ref{table:exp_gate} shows the results of the ablation study of the gating module. The proposed gating module shows the best error rate for three blur conditions than HR-only, random-gating, and iris diameter gating. Our gating module can robustly select a module in REMs with a lightweight network. These results demonstrate both down-sampling and blurring conditions are required to select a module in REMs under various degradation conditions effectively.

\subsection{Comparison}
We compared the performance of the proposed model with conventional methods for different levels of down-sampling and blurring, respectively, to validate the robustness to different levels of resolution degradation. Recognition performances were compared using datasets with fifteen down-sampling levels and six blurring levels. We used CASIA-Iris-Thousand dataset for training. We compare the performance using CASIA-Iris-Thousand to evaluate the recognition performance for a specific dataset. To evaluate the generalization performance, we further evaluated models using three out-of-training test datasets: CASIA-Iris-Lamp, CASIA-Iris-Distance and MMU2.

Figure \ref{fig:compare} (a) shows EER plot for the down-sampled iris images. We evaluate the performance of the iris SR methods, IrisDNet and D-ESRGAN for the iris images with just each model’s target iris diameter (40 pixels or 80 pixels) because they upscale an iris image with a fixed magnification rate. Results of conventional static iris recognition methods show there is a trade-off between high recognition performance for clean HR images and robustness to LR images. Our method overcame this trade-off and achieved the best or almost the best performance for both HR and LR iris images of all four datasets. Figure \ref{fig:compare} (b) shows EER plot for the Gaussian-blurred iris images. The proposed method also achieved the best or almost the best performance for the different blurring conditions for four datasets. These results confirmed that our proposed method enhances the robustness to both down-sampling and Gaussian-blurring for in-training and out-of-training datasets.

We compared the recognition performance on a dataset containing iris images degraded due to both down-sampling and blurring under various conditions. For this evaluation, we made four degraded datasets from four original datasets: CASIA-Iris-Thousand (CASIA-T), CASIA-Iris-Lamp (CASIA-L), CASIA-Iris-Distance (CASIA-D) and MMU2. We applied both down-sampling and Gaussian-blurring operators to all test images. The degradation level was uniformly sampled. The range of a standard deviation ($\sigma$) of GB was (0, 5), and the range of pixels of an iris diameter was (20, 160). The number of images for each degraded dataset was five times that of each original dataset. Table \ref{tab:general} shows an EER for each degraded dataset. The evaluation results show that the proposed method achieved the best performance for all four datasets. These results demonstrate that the proposed method effectively enhanced robustness to various and unknown degradation conditions, even for out-of-training domain datasets.

\begin{table}[tb]
 \caption{Results (EER) for images degraded under various conditions. All methods were trained by CASIA-Iris-Thousand (CASIA-T). Bold is best and underlined is second.}

 \label{tab:general}
 \begin{center}

\scriptsize
  \begin{tabular}{lcccc}
  \hline

    &\multicolumn{4}{c}{Test dataset} \\
   Method        & CASIA-T   & CASIA-L   & CASIA-D   & MMU2 \\        
   \hline 
    Res50-HR  &  0.1060  &  0.1748  &  0.2210  &  0.1201  \\
    Res50+DeblurGAN \cite{lee2021enhanced}  &  0.0426  &  0.0795  &  0.1583  &  0.0658  \\
    Res50-LR  &  0.0151  &  0.0527  &  0.1214  &  0.0433  \\
    Res50-Ring loss \cite{zheng2018ring} &  0.0163  &  0.0538  &  0.1162  &  0.0441  \\
    Res50-KT \cite{boutros2022low} &  \underline{0.0123}  &  \underline{0.0416}  &  \underline{0.0902}  &  \underline{0.0284}  \\
    Res50+Proposed  &  \bf{0.0103}  &  \bf{0.0393}  &  \bf{0.0850}  &  \bf{0.0280}  \\
   \hline
  \end{tabular}
 \end{center}
\end{table}

\begin{figure}[t]
\begin{center}
 \includegraphics[width=1.0\linewidth]{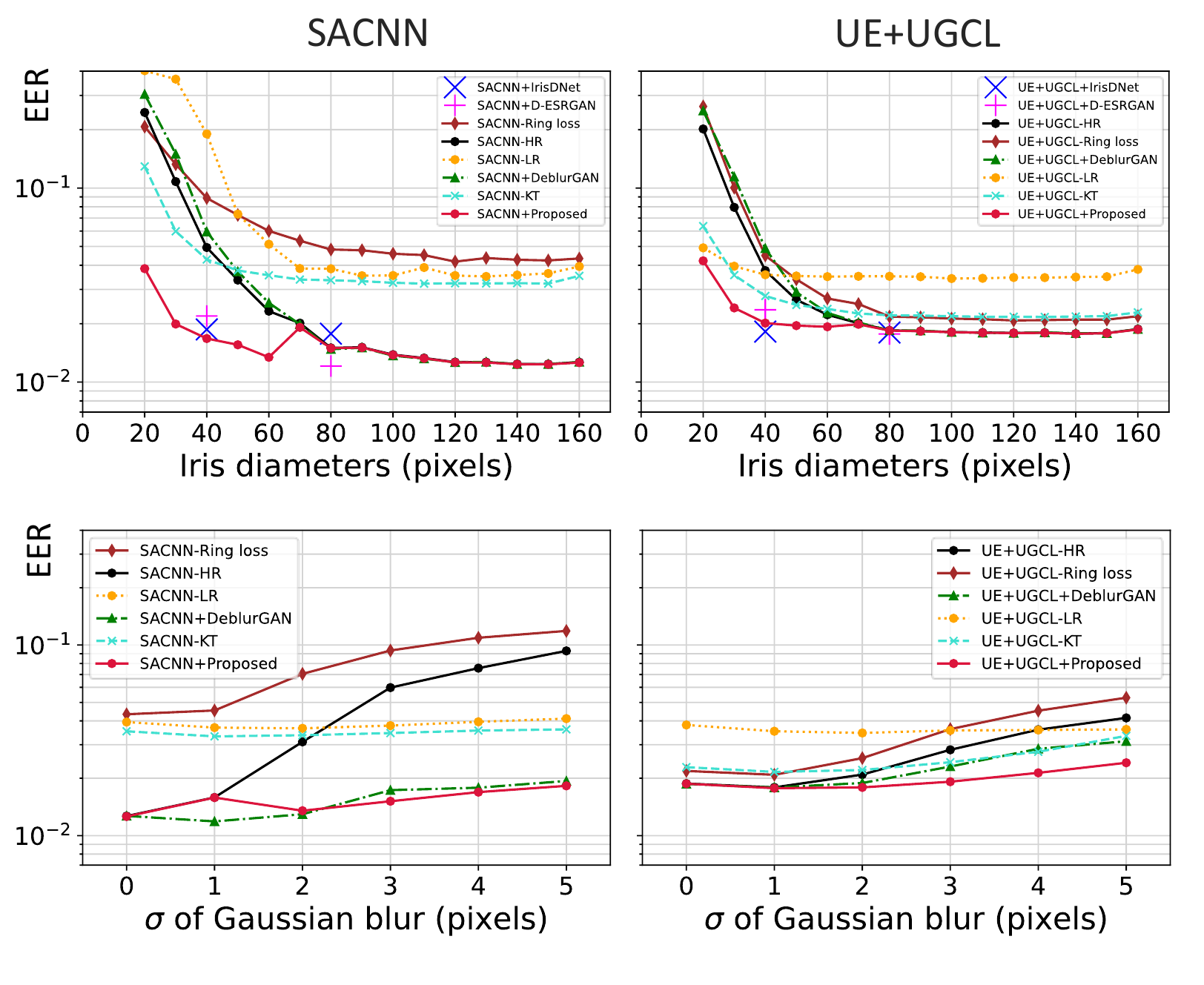}
\end{center}
\caption{Recognition performance comparison using SACNN and UE+UGCL as baseline models. Upper side and lower side shows EER for down-sampled and Gaussian-blurred images.}
\label{fig:compare_backbone}
\end{figure}

We applied our framework to deep-learning-based models for iris recognition and compared our method and conventional methods in addition to the above experiment, where ResNet50 was used as a baseline model. As baseline models, we chose two state-of-the-art metric learning-based methods, SACNN and UE+UGCL. SACNN extracts occlusion-robust features and utilizes spatial information. UE+UGCL extracts features robust to uncertainty degradation factors associated with the image acquisition process. As comparison methods, we trained these baseline models in four ways, as in the case of ResNet50. Detail description of the settings of our method and conventional methods are provided in the supplementary materials. We used CASIA-Iris-Thousand for training and evaluation. Figure \ref{fig:compare_backbone} shows recognition performance comparisons of our method and conventional methods. This result shows our framework can enhance recognition performance under various resolutions with two different baseline models. We evaluated recognition performance using the DET curve or d-prime for a more detailed comparison. Furthermore, we comapred the runtime of our method and conventional methods. These results are provided in the supplementary materials.

\section{Conclusion}
This paper proposed a deep feature extractor for iris recognition at arbitrary resolutions. The method consists of a gating module, three resolution experts and a shared module. The gating module selects the best expert based on the degradation condition of an input image. We also introduced the distillation-based training for LR experts and the shared module to extract common identity features from each expert. We applied our framework to conventional metric learning models. The experimental results showed that our method can improve their recognition performance at various resolutions.

{\small
\bibliographystyle{ieee}
\bibliography{main}
}

\input{X_suppl}
\end{document}

%% file: X_suppl.tex
\clearpage
\setcounter{page}{1}
\maketitlesupplementary

\section{Architecture of the gating module}
\label{sec:gating}
For the proposed dynamic scheme to work in less computation, the gating module needs to work in less computation than feature extraction. To achieve this, we designed the gating module architecture as a lightweight network. To reduce the size of the network, we removed bottleneck block 2 to 4 from ResNet50. Then the gating module consists of an init block, a bottleneck block 1, a global average pooling layer and a fully connected layer of the ResNet50. Its input is a cropped iris image resized to 192$ \times$192 with an iris diameter of 160 pixels.  It takes 5.1 ms for label selection per single image on a CPU (Intel(R) Core(TM) i7-7700 CPU @ 3.60GHz).

\section{Settings of applying our scheme to SACNN and UE+UGCL}
We applied our method to two deep-learning-based iris feature extraction models and evaluated the effectiveness of our framework. As baseline models, we choose simple and accurate CNN (SACNN) based on SE-ResNet18 \cite{kawakami2022simple} and Max-out4 with uncertainty embedding and uncertainty-guided curriculum learning (UE+UGCL) \cite{wei2022towards}. SACNN divides a normalized iris image to four regions, extracts a feature from each divided image, and calculates the occlusion-aware weighted matching score. UE+UGCL extracts an identity feature independent of uncertainty factors. Its model is trained in an easy-to-hard order according to the uncertainty of training images. In the following, we describe the network architecture and training conditions when we apply our framework to these models.

SACNN uses SE-ResNet18 for feature vector extraction and it has an init block, four intermediate blocks and a fully-connected layer as ResNet50. SACNN inputs normalized iris images divided into four regions and their size is 64$\times $128. The dimension of a feature vector is 512. To construct the proposed feature extractor, we split the SACNN trained with HR images at the point between block 2 and block 3 and use the input side as HR-module and the output side the shared-module according to the ablation study of the split point of ResNet50. We add middle-resolution expert module and low-resolution expert module to the model and build REMs. We trained the HR module and shared module in the way proposed in the original paper. The optimization method was Adam \cite{kingma2014adam}, and the initial learning rate was set to 0.0001. The batch size was set to 8. The training was terminated when EER did not improve for 30 epochs. We additionally applied Random Erasing \cite{zhong2020random} to training images as data augmentation. The training settings of the lower-resolution modules are the same as those of ResNet50.

UE+UGCL uses Maxout4 with a Batch normalization and it has four layers to extract maxout-feature-map \cite{wei2022towards, wu2018light} and a head layer that extracts feature vector from the feature map. UE+UGCL inputs normalized iris images preprocessed using their enhancement method. The input size is 128 $\times$ 128 and the dimension of a feature vector is set to 256. We implemented the UE+UGCL model using a GitHub code provided by the author of the original paper \cite{wei2022towards}. To introduce our framework to UE+UGCL, we split the network trained with HR images at the point between layer 2 and layer 3. We regard the input side as HR-module and the output side the shared-module. We add two lower-resolution expert modules to the input side of the network and build REMs. We trained the HR-module and shared-module according to the paper \cite{wei2022towards} and GitHub code. ArcFace loss was used as a classification loss function. The value of margin and scale parameters were set to 0.5 and 64, respectively. The hyper-parameter $\lambda$, a weight of a Kullback-Leibler regularization term, was set to 0.01. The optimization method was SGD with an initial learning rate of 0.001, a momentum value of 0.9, and a weight decay of 0.001. The batch size was set to 300. The total number of epochs was 100 and the learning rate was multiplied by 0.1 for 30, 50 and 70 epochs. The training settings of the lower-resolution modules are the same as those of ResNet50.

\section{Training condition of comparison methods}
\label{sec:comp_set}

As conventional methods, we trained baseline models (Base) in four ways and obtained four models: Base-HR, Base-LR, Base-Ring loss \cite{zheng2018ring} and Base-KT \cite{boutros2022low}. We used ResNet50 (Res50), SACNN and UE+UGCL as the baseline models. To train Base-50-HR and Base-LR, we used the same loss function as used for training the HR and shared modules. We used ring loss to train Base-Ring loss model. We used Softmax cross entropy loss as a classification loss and set the hyper-parameter $\lambda$, a weight of ring loss, to 0.01 according to \cite{zheng2018ring}. The Base-KT model was trained using the loss function for KT proposed for LR iris recognition \cite{boutros2022low}. We used the same data-augmentation methods as used for the HR and share module training. We additionally applied down-sampling and Gaussian-blurring to iris images to train Base-LR, Base-Ring loss and Base-KT. The optimization settings of the baseline models were the same as those for the HR and shared modules.

\begin{figure*}[t]
\centering
  \includegraphics[width=0.99\linewidth]{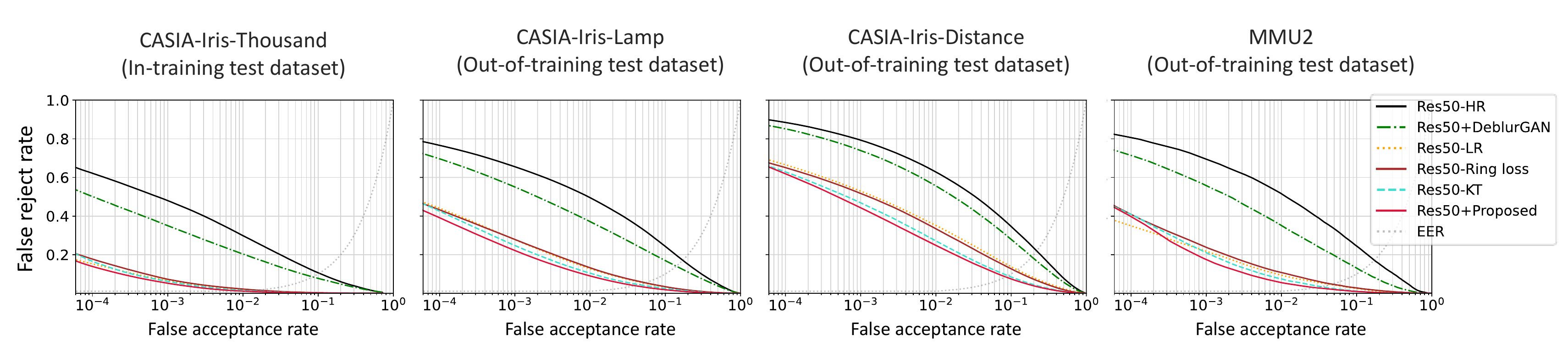}
   \caption{DET curve comparison for iris images degraded under various conditions.}
\label{fig:det_res50}
\end{figure*}

As image-restoration-based methods, we trained IrisDNet \cite{IrisDNet}, D-ESRGAN \cite{wang2022d} and DeblurGAN \cite{kupyn2018deblurgan, lee2021enhanced}. We trained IrisDNet model with a two-magnification rate and the model with a four-magnification rate. Each model takes an iris image with an iris diameter of 80 pixels and 40 pixels as an input, respectively. We calculated the identity preserving loss proposed in \cite{IrisDNet} using RN50-HR, which was used as an iris feature extractor. We trained D-ESRGAN model with a two-magnification rate and the model with a four-magnification rate. We followed the three-stage training scheme and trained the model for three epochs for the first stage, four epochs for the second stage, and forty-three epochs for the third stage. We trained DeblurGAN to reproduce a clean iris image from a blurred iris image. In conventional works \cite{kupyn2018deblurgan, lee2021enhanced}, DeblurGAN was trained to reproduce clean images from images degraded by motion blur kernels. In our experiment, we evaluated the recognition performance for iris images degraded by Gaussian blur. To train DeblurGAN to reproduce clean iris images from degraded iris images by Gaussian blurring, we produced degraded iris images by applying the Gaussian blurring kernel and used them for training. The range of standard deviation of Gaussian-blurring was set to (0, 5). The architecture of DeblurGAN was the same as \cite{lee2021enhanced} and set the value of the input channel to one for a NIR iris image. Also, the hyper-parameter setting for optimization was the same as \cite{lee2021enhanced}.

\begin{table}[tb]
 \caption{D-prime for images degraded under various conditions. All methods were trained by CASIA-Iris-Thousand (CASIA-T) and evaluated using CASIA-T, CASIA-Iris-Lamp (CASIA-L), CASIA-Iris-Distance (CASIA-D) and MMU2. Bold and underlined mean the best and second, respectively.}

 \label{tab:dprime_res50}
 \begin{center}

\scriptsize
  \begin{tabular}{lcccc}
  \hline

    &\multicolumn{4}{c}{Test dataset} \\
   Method        & CASIA-T   & CASIA-L   & CASIA-D   & MMU2 \\        
   \hline 
    Res50-HR  &  2.609  &  1.968  &  1.600  &   1.971  \\
    Res50+DeblurGAN \cite{lee2021enhanced}  & 2.989  &  2.245  & 1.815  & 2.505  \\
    Res50-LR  & 5.143  & 3.382  &  2.402  &  3.131  \\
    Res50-Ring loss \cite{zheng2018ring} &  5.086  &  3.265  &  2.453  &  2.935  \\
    Res50-KT \cite{boutros2022low} &  \underline{5.415}  &  \underline{3.559}  &  \underline{2.789}  &  \underline{3.538}  \\
    Res50+Proposed  &  \bf{5.514}  &  \bf{3.769}  &  \bf{2.893}  &  \bf{3.809}  \\
   \hline
  \end{tabular}
 \end{center}
\end{table}

\section{Results for detail comparison}
We drew detection error trade-off (DET) curve and evaluated d-prime to compare our method and conventional methods and validated the effectiveness of our method. DET curve illustrates the false reject rate at the different false acceptance rates. D-prime measures the degree of separation of matching score distribution of the same eye and that of a different eye. D-prime is calculated from means and standard deviations from two distributions.
\begin{equation}
	d' = \frac{|\mu_{1}-\mu_{2}|}{\sqrt{(\sigma_{1}^2+\sigma_{2}^2)/2}},
	\label{eq:dprime}
\end{equation}

Figure \ref{fig:det_res50} shows DET curve for iris images degraded due to both down-sampling and blurring under various conditions. We used ResNet50 as a baseline model and used CASIA-Iris-Thousand for training. We used four test datasets: CASIA-Iris-Thousand (CASIA-T), CASIA-Iris-Lamp (CASIA-L), CASIA-Iris-Distance (CASIA-D) and MMU2. They are used for evaluation of table \ref{tab:general}. This result shows our method enhances the false reject rate at a low false accept rate, such as 0.001 and 0.0001. Table \ref{tab:dprime_res50} shows d-prime for each dataset. The proposed method shows the best d-prime performance, even for out-of-training datasets. These results demonstrate that the proposed method can effectively improve the recognition performance under various resolutions.

In addition to the evaluation with publicly available datasets, we collected paired multiple-resolution iris images to verify our claims using real-world LR images. We captured the images of 20 irises from 10 subjects using a 35-mm lens, a camera device (Baumer vcxu-65M.R) and a NIR lighting device with an 850-nm wavelength. The distance between the camera and the subject was adjusted so that the iris diameter was 160, 120, 80, and 40 pixels. The number of images was 50 per iris at one position, for a total of 4000. We used HR iris images (160 pixels) as gallery images and evaluated the error rate for each LR iris image dataset and all lower-resolution (All LR) iris images. We compared recognition performance with our method and conventional methods using ResNet50 as a baseline. We trained all models with CASIA-Iris-Thousand. Table \ref{table:exp_realLR} shows the EER for each resolution. This result demonstrates the effectiveness of the proposed method for real-world LR iris images.
Note that we collected image data after obtaining approval regarding ethics from our affiliated organization.

\begin{figure}[t]
\centering
  \includegraphics[width=0.99\linewidth]{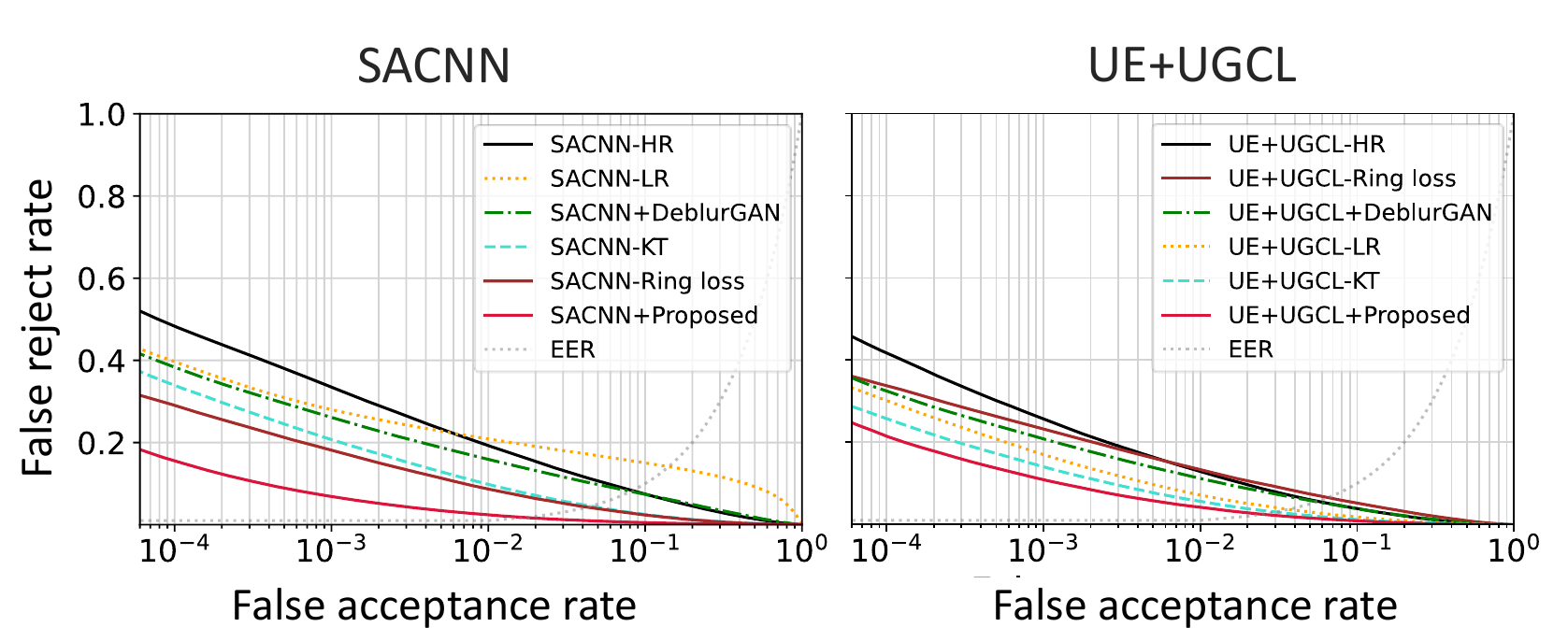}

   \caption{DET curve comparison for iris images degraded under various conditions using SACNN and UE+UGCL as a baseline models.}
\label{fig:det_sacnn_ueugcl}
\end{figure}

\begin{table}[tb]
 \caption{EER and d-prime for iris images degraded various resolution. Bold is best and underlined is second.}

 \label{tab:eer_dprime_sacnn}
 \begin{center}

  \begin{tabular}{lcccc}
  \hline

   Method        & EER   & d-prime\\        
   \hline 
    SACNN-HR  &  0.0800  & 1.894   \\
    SACNN+DeblurGAN \cite{lee2021enhanced}  & 0.0774  &  2.077   \\
    SACNN-LR  & 0.1453  & 2.066    \\
    SACNN-Ring loss \cite{zheng2018ring} & \underline{0.0436}  &  \bf{3.470}   \\
    SACNN-KT \cite{boutros2022low} &  0.0454  &  3.002  \\
    SACNN+Proposed  &  \bf{0.0177}  &  \underline{3.084}   \\
    \hline
    UE+UGCL-HR  & 0.0562  &  3.118   \\
    UE+UGCL+DeblurGAN \cite{lee2021enhanced}  & 0.0560  &  3.286   \\
    UE+UGCL-LR  & 0.0371  & 3.796   \\
    UE+UGCL-Ring loss \cite{zheng2018ring} &  0.0644  &  2.678   \\
    UE+UGCL-KT \cite{boutros2022low} &  \underline{0.0304}  &  \bf{4.152}  \\
    UE+UGCL+Proposed  &  \bf{0.0249}  &  \underline{4.133}   \\
   \hline
  \end{tabular}
 \end{center}
\end{table}

\begin{table}[t]
  \scriptsize
  \caption{Evaluation results (EER) for real world LR iris images using optically captured multiple-resolution paired iris images. Bold is best and underlined is second.}
  
   \label{table:exp_realLR}
   \centering
    \small
    \begin{tabular}{lcccc}\hline
     &\multicolumn{3}{c}{Iris diameter [pixels]} \\
     Method      &    120 &   80  &    40 &    All LR   \\ 
     \hline 
        D-ESRGAN \cite{wang2022d} &  -  &  0.0743  &  -  &  0.1725  \\
        IrisDNet \cite{IrisDNet}   &  -  &  0.1034  &  -  &  0.1265  \\
        RN50-HR  &  0.0365  &  0.0767  &  0.2302  &  0.1635  \\
        DeblurGAN \cite{lee2021enhanced} &  0.0329  &  0.0763  &  0.2193  &  0.1547  \\
        RN50-LR  &  0.0346  &  \underline{0.0660}  &  0.1503  &  0.9893  \\
        Ringloss \cite{zheng2018ring} &  0.0356  &  0.0879  &  \underline{0.1419}  &  0.0944  \\
        KT \cite{boutros2022low} &  0.0347  &  0.0736  &  0.1538  &  \underline{0.0905}  \\
        Proposed  &  \bf{0.0130}  &  \bf{0.0381}  &  \bf{0.1278}  &  \bf{0.0749}  \\
     \hline
  \end{tabular}
  \end{table}

\section{Runtime comparison}
 We compared the runtime for a single image of the proposed and that of conventional methods on a CPU (Intel(R) Core(TM) i7-7700 CPU @ 3.60GHz). Localization process by ILN takes 19.5 ms. Runtime of normalization process for ResNet50, SACNN and UE+UGCL are 2.1 ms, 2.2 ms, and 2.3 ms, respectively. Runtime of static feature extraction process for ResNet50, SACNN and UE+UGCL are 67.9 ms, 50.1 ms, and 32.2 ms. IrisDNet and D-ESRGAN with four magnification rates take 458.4 ms and 1006.2 ms for single image SR, respectively. DeblurGAN takes 612.4 ms for a single image deblurring. The proposed method requires only 5.1 ms additionally for gating module calculation. These results demonstrate that the proposed method expands robustness with less computational cost than the image-restoration-based methods.

\section{Iris localization performance for LR images}
\begin{figure}[t]
\centering
  \includegraphics[width=0.85\linewidth]{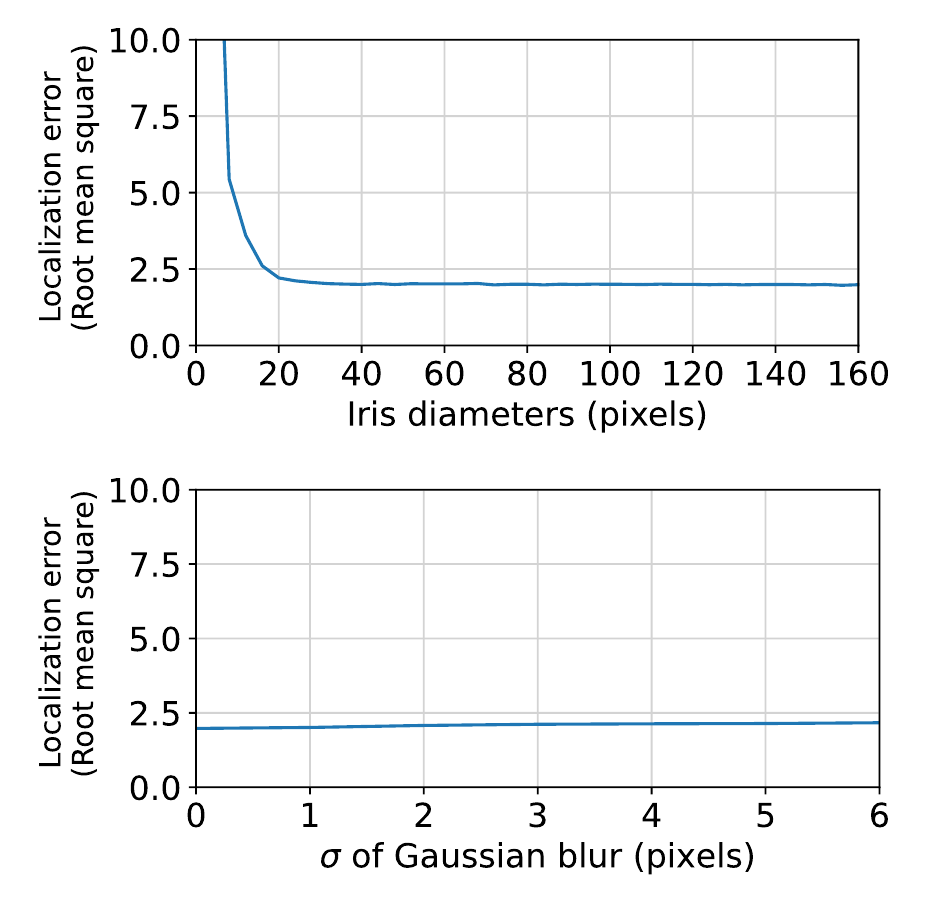}

   \caption{Iris localization performance for down-sampled iris images and Gaussian-blurred iris images.}
\label{fig:ILN_LR}
\end{figure}
We located the center coordinates and radii based on the HR iris images and used them even for degraded iris images for a fair comparison of feature extraction methods. However, resolution degradation also may affects the localization performance. To validate the feasibility of our localization settings, we evaluate the localization performance of Iris localization networks (ILN) \cite{Toizumi_2023_WACV} using down-sampled iris images and Gaussian-blurred iris images. We used CASIA-Iris-Thousand for the test dataset. Figure \ref{fig:ILN_LR} shows the error value for each degradation condition. The error value is the root mean square of the center coordinates and radii of the iris and pupil circles for iris images. Upper side of the figure shows that localization error is maintained even for iris images with a diameter of twenty pixels. Lower side of the figure shows that localization error is maintained for five pixels of standard deviation of Gaussian-blurring. These results indicate the localization performance of ILN is more robust to resolution degradation than feature extraction methods.